\documentclass[runningheads]{llncs}
\usepackage{xcolor}

\usepackage[T1]{fontenc}
\usepackage{graphicx}
\begin{document}
%
\title{Using Backbone Foundation Model for Evaluating Fairness in Chest Radiography Without Demographic Data}
\titlerunning{Evaluating Fairness in Chest Radiography Without Demographic Data}
%
 
\author{\index{Queiroz, Dilermando} Dilermando Queiroz\inst{1} \and
\index{Anjos, André}André Anjos\inst{2} \and
\index{Berton, Lilian}
Lilian Berton\inst{1}}

\authorrunning{D. Queiroz et al.}
%
\institute{Universidade Federal de São Paulo, São Paulo, Brazil \and
Idiap Research Institute, Martigny, Switzerland\\
}
\maketitle              
\begin{abstract}
Ensuring consistent performance across diverse populations and incorporating fairness into machine learning models are crucial for advancing medical image diagnostics and promoting equitable healthcare. However, many databases do not provide protected attributes or contain unbalanced representations of demographic groups, complicating the evaluation of model performance across different demographics and the application of bias mitigation techniques that rely on these attributes. This study aims to investigate the effectiveness of using the backbone of Foundation Models as an embedding extractor for creating groups that represent protected attributes, such as gender and age. We propose utilizing these groups in different stages of bias mitigation, including pre-processing, in-processing, and evaluation. Using databases in and out-of-distribution scenarios, it is possible to identify that the method can create groups that represent gender in both databases and reduce in 4.44\% the difference between the gender attribute in-distribution and 6.16\% in out-of-distribution. However, the model lacks robustness in handling age attributes, underscoring the need for more fundamentally fair and robust Foundation models. These findings suggest a role in promoting fairness assessment in scenarios where we lack knowledge of attributes, contributing to the development of more equitable medical diagnostics.

\keywords{Fairness  \and Medical Image \and Foundation Model}
\end{abstract}
\begin{figure}
\includegraphics[width=\textwidth]{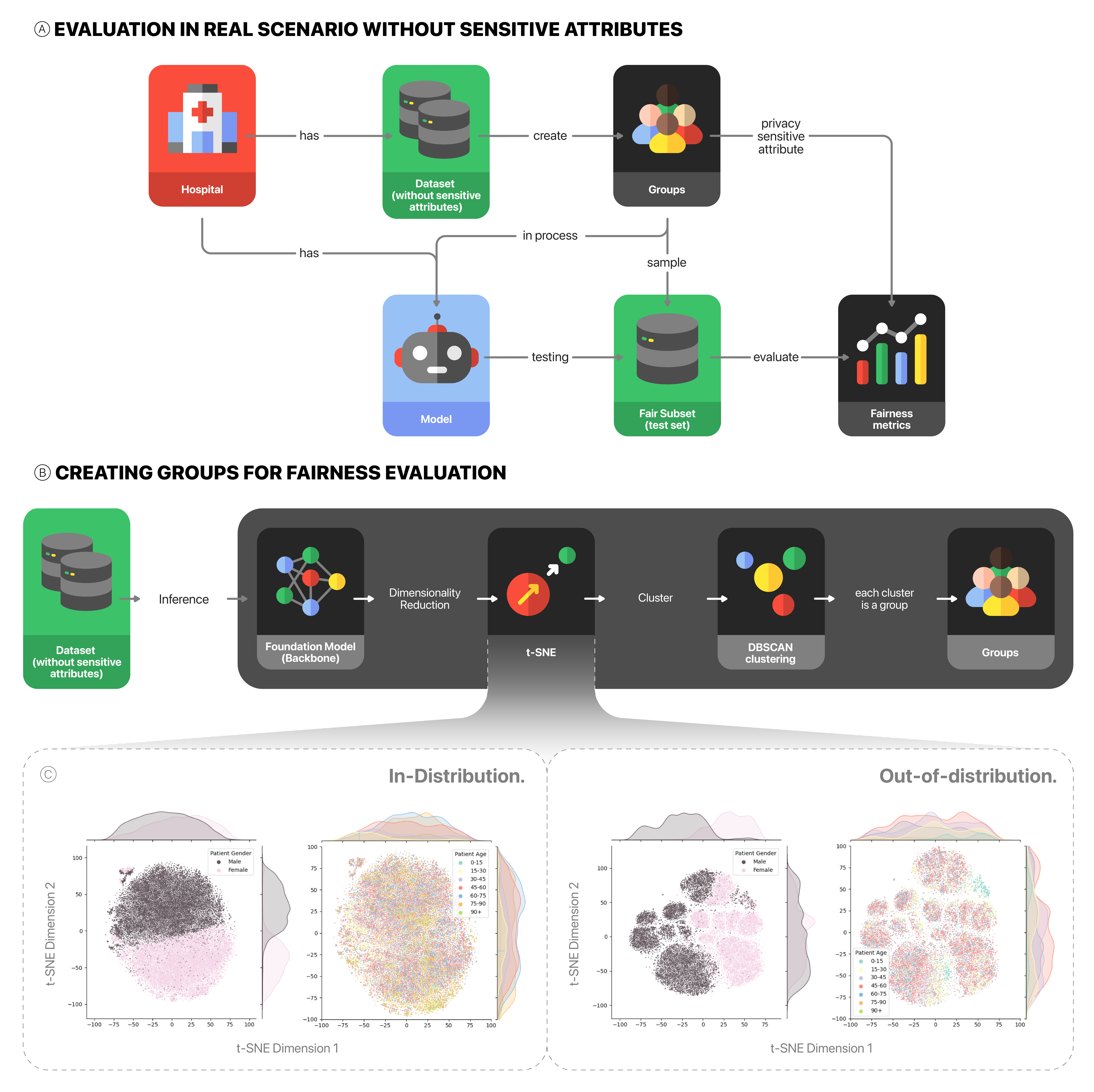}
\caption{\textbf{(a)} Overview of the application of groups formed by the proposed method in various contexts such as model processing, subset selection, and metric evaluation.
\textbf{(b)} The process begins with a Foundation Model (FM), trained on a large corpus of chest X-ray images, to extract embeddings from a dataset devoid of sensitive attributes. These embeddings are then subjected to dimensionality reduction via t-SNE~\cite{maaten_visualizing_2008}, facilitating clustering in a lower-dimensional space and enhancing computational efficiency. Subsequently, DBSCAN \cite{ester_density-based_1996} is applied to identify clusters that will be used to form a notion of groups.
\textbf{(c)} Visualization of embeddings, which were subsequently reduced to two dimensions using t-SNE. These dimensions are denoted by patient age and gender, spanning across the CheXpert (in-distribution) and NIH (out-of-distribution) databases.
} \label{fig:method}
\end{figure}

\section{Introduction}
Recent advancements in medical diagnosis, particularly through Deep Learning (DL) techniques and cloud computing, have the potential to enhance diagnostic accuracy and accessibility. For example, cloud-based DL systems 
can streamline diagnostics across hospitals, providing crucial tools for medical professionals. However, the rapid proliferation of DL algorithms in healthcare raises ethical concerns about their impact on underrepresented communities \cite{mccradden_ethical_2020}. Studies show that Artificial Intelligence (AI) can identify causal structures in data correlated with protected characteristics such as race, gender, age, and ethnicity \cite{glocker_algorithmic_2023}, potentially exacerbating healthcare inequalities by using these correlations to predict health outcomes.

A common issue with medical image data is that many publicly available datasets do not provide demographic information. For example, in chest X-ray datasets, only a few include protected attributes \cite{gichoya_ai_2022}, making it challenging to evaluate DL models trained on such data across different demographics and sensitive variables. Moreover, the vast majority of fairness techniques require datasets containing this information \cite{chen_algorithmic_2023}. These methods span pre-processing, in-processing, and post-processing stages and are recommended by global health authorities like the World Health Organization to foster equity in healthcare. Therefore, techniques to mitigate unfair treatment from DL models that do not rely on these attributes are essential for the development of the field.

The introduction of new techniques, such as self-supervised learning (SSL), represents a significant leap forward. The use of more images without the need for specific labels using SSL in healthcare allows Foundation Models (FM) to be trained with large quantities of unannotated data, bypassing expensive and tedious labeling processes. An FM is a model trained on extensive and diverse data, typically employing self-supervision at scale, which can then be adapted or fine-tuned for a variety of downstream tasks. Furthermore, FM can help in building more robust models that can be used in a variety of distribution data \cite{azizi_robust_2023}.

The contributions of this work are delineated as follows:
\begin{itemize}
    \item We leverage the backbone of an FM to construct groups that approximate sensitive attributes, facilitating fairness evaluation methods for datasets without demographic data.
    \item We propose a comprehensive evaluation and bias mitigation framework tailored for contexts lacking demographic attributes in medical images.
    \item We show that the FM used in the framework is more robust for gender than for age in demographic data.
\end{itemize}



\section{Related work}
In recent years, there has been an increasing amount of literature on fairness without demographic bias. A notable contribution in this area categorizes methods for achieving fairness without demographic data into four groups: collecting demographic data, implementing additional protections for data collection, utilizing auxiliary datasets and inferring demographic data, and exploring alternatives to traditional group fairness approaches \cite{ashurst_fairness_2023}. These methods aim to address the challenges posed by the lack of accurate, complete, or available demographic information, thereby providing a comprehensive framework for researchers, practitioners, and policymakers to navigate the complex landscape of algorithmic fairness. Our method aligns with the category of inferring demographic data. However, these methods exhibit variable performance that can disproportionately impact already marginalized groups \cite{scheuerman_how_2020,scheuerman_how_2019}. We propose using the Foundation Model to extract image embeddings without inferring protected attributes. Due to self-supervised learning, these models are trained solely without labels and can reduce bias \cite{vaidya_demographic_2024}.

Our proposed method exhibits similarities with the cluster-based balancing approach referred to as fair class balancing \cite{yan_fair_2020}. Both methodologies exploit the intrinsic group structure present within the data, identifying naturally occurring homogeneous subgroups characterized by shared feature similarities through clustering in the feature space. By employing these clusters, our method ensures that the training process adequately represents the diversity within the dataset without directly predicting sensitive attributes. We aim to demonstrate that these groups are representative and can be used for fairness evaluation and other applications.

A similar approach incorporates clustering to enhance model fairness and robustness in end-to-end speech recognition systems \cite{veliche_improving_2023}. While both workflows employ clustering strategies, our method focuses on image data rather than speech data. In their approach, embeddings are used to create clusters, which are utilized during model training to address privacy concerns and improve fairness. Similarly, our method aims to enhance fairness without directly using protected attributes, highlighting the versatility and applicability of clustering techniques across different data domains for promoting fairness and preserving privacy.



\section{Materials and Methods}

In this section, we describe the methodology used to conduct our research. The real scenario process is illustrated in Figure 1a, and Figure 1b shows a detailed flowchart of each step proposed in the work. In the following paragraphs, we detail the method.

\textbf{\textit{Datasets}} For the analysis of the framework, two radiology image datasets were employed: CheXpert \cite{johnson_mimic-cxr_2019} and ChestX-ray14 (US National Institutes of Health (NIH)) \cite{wang_chestx-ray8_2017}. We chose CheXpert as the in-distribution (ID) dataset, as the base model of the FM was pre-trained using this set. To evaluate the generalization capability of the technique across different datasets, we adopted NIH as an out-of-distribution (OOD) dataset, thereby ensuring the robustness of the approach.

CheXpert contains 224,316 chest radiographs from 65,240 patients, sourced from Stanford Hospital (2002-2017). Only one image per patient was selected, focusing on those with the five most common pathologies (atelectasis, consolidation, pulmonary edema, pleural effusion, and cardiomegaly), resulting in 58,662 images. The NIH dataset includes 112,120 annotated X-rays from 30,805 patients. Similarly, one image per patient was chosen, yielding 30,802 images.




Both datasets provide metadata on age and gender. CheXpert has 55.4\% male and 44.6\% female patients, while NIH has 53.9\% male and 46.1\% female. Age distributions differ, with NIH being more unbalanced. Most samples in both datasets are from patients aged 45-65 years, with CheXpert having more older individuals and limited samples from younger individuals. The FM was trained on CheXpert, which lacks pediatric data, whereas NIH includes this age group.

The datasets are the first component of the methodoly that as shown in Figure \ref{fig:method}

\textbf{\textit{Extract embedings}} The chosen FM for the method was REMEDIS \cite{azizi_robust_2023}, with input images of 448x448 pixels and three channels, trained from the BiT-M \cite{kolesnikov_big_2020} backbone. According to the description, the model underwent initial pre-training on an extensive set of natural images, followed by a second phase of pre-training using self-supervised learning. The specific technique used for pre-training and learning representations is SimCLR \cite{chen_simple_2020}.

We utilized a REMEDIS pre-trained backbone to extract embeddings from selected images in both datasets. This backbone was employed without further training and used solely for image inference. To address the challenge of visualizing and the computational cost of high-dimensional embeddings, we employed the t-SNE method \cite{maaten_visualizing_2008}. This technique reduces the embeddings to two dimensions, making visualization possible.



\textbf{\textit{Clusters}} Previous research has established some notion of fairness \cite{chen_algorithmic_2023}, the three commonly used fairness criteria for binary classification tasks are demographic parity, predictive parity, and equalized odds. These definitions establish a set of $(X, Y, A)$ where $X$ are the samples, $Y$ are the labels, and $A$ are the protected attributes. However, using these metrics is challenging in databases without protected attributes.

To address this, we categorized the data into sets representing specific image characteristics after extracting and reducing the dimensionality of the embeddings. We used DBSCAN \cite{ester_density-based_1996} to group images by similar characteristics in feature space. This approach allows each image to be assigned to a cluster representing a protected attribute, thus forming the set $A$ using cluster IDs.

The number of clusters directly impacts their size: fewer clusters create larger, more generalized groups, while more clusters result in smaller, more specific groups. We chose to use 15 to 25 clusters to create medium-sized groups. Accordingly, we adjusted the DBSCAN \cite{ester_density-based_1996} parameters to achieve this range while minimizing unclustered data labeled as $-1$.

In this approach, we set the minimum number of samples required for each cluster to 120 for the CheXpert dataset and 40 for the NIH dataset. The maximum distance between two samples was set to 4 for CheXpert and 3 for NIH. We determined the number of clusters to be 15 for CheXpert and 22 for NIH, with 14,785 unclustered data points for CheXpert and 2,879 for NIH. To balance the database, we utilized the previously defined clusters and sampled an equal number of samples from each cluster to reach 30\% of the original dataset. Since each cluster represents a set of protected attributes, this approach allows us to have a more representative sample of our database.

\section{Results and Discussion}

\begin{figure}
\includegraphics[width=\textwidth]{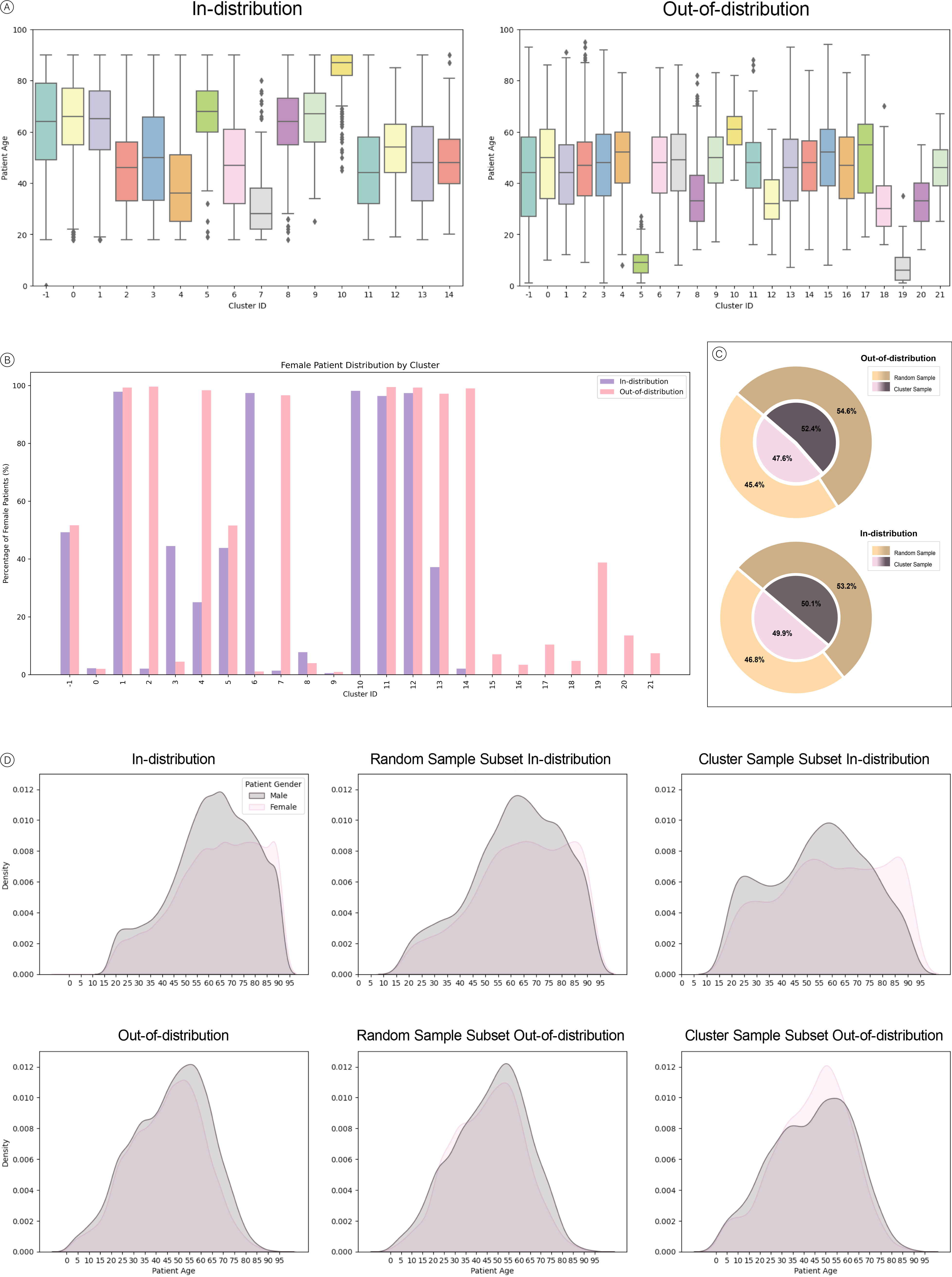}
\caption{\textbf{(a)} Age distribution across clusters in the CheXpert (in-distribution) and NIH databases (out-of-distribution). The cluster $-1$ represents the unclustered data; other numbers are the clusters. \textbf{(b)} Gender Distribution Across Clusters for CheXpert and NIH Datasets. \textbf{(c)} Gender Distribution after sampling 30\% of CheXpert and NIH Datasets.
\textbf{(d)} Kernel density estimate (KDE) plots illustrate the comparison of age distributions between the CheXpert and NIH subsets, utilizing both random and cluster sampling techniques, with data categorized by patient gender.
}
\label{fig:results}
\end{figure}

\textbf{\textit{Feature Space Analysis}} The first set of analyses examined the proposed method's impact on evaluating fairness in datasets without protected attributes. We analyze the feature spaces of the FM using the metadata about age and gender from both databases, and these attributes are used for visualization and verification of the representative of each generated group. Figure \ref{fig:method}c shows the embeddings after dimensional reduction to two dimensions, facilitating the visualization of images extracted by the backbone of the FM. We can visualize strong separation based on the gender attribute in both databases. 

However, the distinction based on age is not as clear in both databases. A benchmark study comparing methods for improving fairness using a supervised learning approach \cite{zong_medfair_2023} shows that the methods yield better results for gender than age across different approaches. As demonstrated in Figure \ref{fig:method}c, self-supervised learning exhibits the same issue. These results suggest that models trained on radiography images generally have more difficulty separating the age attribute in the feature space than the gender attribute. As shown in Figure \ref{fig:method}c in the NIH database, the $0-15$ years is the only group clearly defined. The other groups are more mixed, making it more challenging to visualize a separation.

\textbf{\textit{Age and Gender Representation in Clusters}} Figure \ref{fig:results}a compares the age distribution across clusters for both datasets. Clusters representing more common age groups are more prevalent than those for less common age groups. In the in-distribution dataset, we have clusters represented by different age distributions. However, some clusters share the same distribution. Interestingly, clusters with the same age distribution may correspond to distinctly different groups concerning protected attributes. For instance, clusters 0 and 1 exhibit similar age profiles but differ significantly in gender composition. Cluster 0 contains only 2.20\% females, in contrast to cluster 1, which is composed of 97.86\% females as shown in Figure \ref{fig:results}b. This gender metadata can be more distinctly segregated, as shown by the clusters for both datasets, where we have clusters with an equal proportion of males and females and clusters composed entirely of one gender.

It is noticeable that within the CheXpert dataset, only one cluster represents individuals aged over 80 years, and similarly, only one cluster for those under 40 years. In contrast, most clusters correspond to more prevalent represented data in the out-of-distribution dataset. Only two groups, clusters 5 and 19, define underrepresented data. Surprisingly, the age range of these clusters is between 0 and 20 years, a demographic not present in the in-distribution dataset, indicating that the model has not encountered this data during training.

These groups must be used with caution because they may contain biases towards certain groups. As they are not exclusively defined by a single protected attribute, there may be small, nested groups that are overlooked in the evaluation process. For instance, in Figure \ref{fig:results}a, within the out-of-distribution scenario, there is an absence of a group representing individuals over the age of 70, likely because they are included within other overlapping groups.

\textbf{\textit{Generating Balanced Subsets}} We leveraged the fact that these groups might represent protected variables, thus enabling us to sample an equal number of individuals from each group. With this approach, groups more prevalent in the dataset are combined into the same cluster, reducing the likelihood of these groups being over-sampled. Figure \ref{fig:results}d compares two techniques: random sampling and the cluster sampling method proposed in this paper. We utilized the kernel density estimate (KDE) plot to visualize the distribution of the patient age and patient gender in subsets sampled from the original database.

As demonstrated in Figure \ref{fig:results}d, the cluster sampling method yields favorable results for the CheXpert database, which contains in-distribution data. The figure shows a clear trend of a decrease in the majority group regarding age and sex in the cluster sample, resulting in a more balanced sampling compared to the random approach. The random sampling exhibits an imbalance that mirrors the original database regarding age and gender.

A plausible explanation for these observations is that groups with a higher probability of occurrence, such as the age range of 55 to 85 years, are grouped into a single cluster or multiple clusters, as illustrated in Figure \ref{fig:results}a. This confirms that these groups can be used to represent protected attributes and, consequently, be sampled to produce a more balanced subset.

However, the results are insignificant for the NIH database, which contains out-of-distribution data. Figure \ref{fig:results}a shows a minor increase in younger age groups and a significant increase in gender distribution. In this case, the random sampling again closely resembles the distribution of the dataset, as expected.

\textbf{\textit{Gender Balance in Sub-Datasets}} The results regarding the gender proportion in the subsets are presented in Table \ref{fig:results}c. In contrast, the cluster sample reduces this disparity, approaching a more balanced distribution of nearly 50\%. Specifically for the CheXpert database, the new subset created through cluster sampling exhibits only a 4.72\% difference between female and male representation. In comparison, this difference is 9.16\% in the random sample, indicating a significant improvement of 4.44\% towards gender balance with the cluster sampling approach.

Using random sampling on the NIH dataset, we achieved a subset with a gender difference of 6.4\%. However, by employing the technique proposed in the article, we reduced this difference to a mere 0.24\%, resulting in an improvement of 6.16\%. This method demonstrates robustness for out-of-distribution data in terms of gender attributes. However, it yields slight improvement for the age attribute.

\textbf{\textit{Robustness Across Distribution Shifts}} An open question in the literature is whether there are methods that ensure the transfer of fairness across distribution shifts \cite{schrouff_diagnosing_2023}. This study demonstrates the feasibility of maintaining fairness for gender across different distributions using the proposed method. Such an outcome is attributable to using FM as an embedding extractor. However, the method needs to prove more efficient for the age attribute in out-of-distribution.




\section{Conclusion}

The study demonstrates a novel approach to promoting fairness in medical image diagnostics, especially when demographic data are unavailable. By leveraging the backbone of Foundation Models to create groups representing protected attributes like gender and age, we can apply mitigation techniques across pre-processing, in-processing, and evaluation stages. The results show significant improvements in gender fairness, with a 4.44\% and 6.16\% reduction in gender attribute disparity for in-distribution and out-of-distribution data, respectively. However, the model faces challenges with age-related attributes, suggesting a need for further development in this area. This research underscores the potential of FMs to advance equitable healthcare diagnostics by providing a framework for fairness evaluation even in the absence of explicit demographic metadata.

\begin{credits}
\subsubsection{\ackname} 
We thanks Fundação de Amparo à Pesquisa do Estado de São Paulo (FAPESP), grants 21/14725-3 and 23/12493-3, Conselho Nacional de Desenvolvimento Científico e Tecnológico (CNPq), Swiss National Science Foundation (SNSF) under Grant No. 200021E\_214653. In addition, we would like to especially thank Lucas Tosta for his assistance in designing the figures.

\end{credits}
%
%
%
\bibliographystyle{splncs04}
\bibliography{fairness-cluster-remedis}

\end{document}